\documentclass[conference]{IEEEtran}

\ifCLASSINFOpdf
\else
\fi

\usepackage{graphicx}
\usepackage[dvipsnames]{xcolor}
\usepackage{amsthm} 
\theoremstyle{definition}
\newtheorem{definition}{Definition}

\begin{document}

\title{SCR-Apriori for Mining `Sets of Contrasting Rules'}

\author{\IEEEauthorblockN{Marharyta Aleksandrova}
\IEEEauthorblockA{
University of Luxembourg,\\
\\
Esch-sur-Alzette, Luxembourg,\\
Email: marharyta.aleksandrova@gmail.com}
\and
\IEEEauthorblockN{Oleg Chertov}
\IEEEauthorblockA{
National Technical University of Ukraine \\
`Igor Sikorsky Kyiv Polytechnic Institute',\\
Kyiv, Ukraine,\\
Email: chertov@i.ua}
}

\maketitle

\begin{abstract}
In this paper, we propose an efficient algorithm for mining novel `Set of Contrasting Rules'-pattern (SCR-pattern), which consists of several association rules. This pattern is of high interest due to the guaranteed quality of the rules forming it and its ability to discover useful knowledge. However, SCR-pattern has no efficient mining algorithm. We propose SCR-Apriori algorithm, which results in the same set of SCR-patterns as the state-of-the-art approache, but is less computationally expensive. We also show experimentally that by incorporating the knowledge about the pattern structure into Apriori algorithm, SCR-Apriori can significantly prune the search space of frequent itemsets to be analysed.
\end{abstract}

\IEEEpeerreviewmaketitle

\section{Introduction}

Association rules learning is a popular technique in data mining \cite{kotsiantis2006association}. However, it is known that finding rules of high quality is not always an easy task \cite{lenca2008selecting}. This issue is even more significant in domains where the reliability of the obtained knowledge is required to be high (for example, in medicine). Also, association rules mining techniques usually generate a huge number of rules that have to be analysed by a human in order to choose meaningful and useful ones \cite{techapichetvanich2004visual}.

Recently a novel pattern called `Sets of Contrasting Rules' was proposed \cite{aleksandrova2016setsecai}. This pattern was proven to produce rules of high-quality \cite{aleksandrova2017contrast}. It was also shown that its structure allows identifying meaningful knowledge, in particular, \textit{local differences} between subgroups of the dataset and \textit{trigger factors}. However, no efficient algorithm for mining SCR-patterns was proposed so far. The state-of-the-art procedure of constructing SCR-patterns performs filtering of association rules obtained by other techniques and chooses those that can form this pattern \cite{aleksandrova2016sets}. Although this approach allows discovering all SCR-patterns in the dataset, it has high level of redundancy, as a vast majority of mined rules is filtered out.

In this paper, we address a question of efficient mining of SCR-patterns. We propose a novel SCR-Apriori algorithm that prunes the search space via incorporating knowledge about the structure of the pattern. The proposed algorithm is a modified version of existing Apriori \cite{agrawal1993mining} and CAR-Apriori \cite{ma1998integrating} algorithms and incorporates advantages of both of them.

\section{Association Patterns}

\subsection{Association Rules}

The core notion of association patterns mining is \textit{item}. Item is defined as a value of a particular attribute in a given record. A set of items $X$ is referred to as an \textit{itemset} and an \textit{association rule} is defined as an induction rule of the form $X\rightarrow Y$, where $X$ and  $Y$ are itemsets and $X \cap Y = \emptyset$ \cite{agrawal1993mining}. The left-hand side of the association rule $X$ is called \textit{antecedent}, and its right-hand side $Y$ is called \textit{consequent}.   

The \textit{support number} of an itemset $X$ on the dataset $D$ is defined as a number of records in $D$ that contain $X$. Similarly, the \textit{support number} of an association rule $X\rightarrow Y$ is defined as a support number of the union $X \cup Y$. Often an alternative measure called \textit{support} is used. It is defined as the ratio of the support number to the total number of records in the dataset. The ratio of the support of the rule to the support of its antecedent defines the \textit{confidence} of the rule on a particular dataset, that is $conf_D\left(X\rightarrow Y\right) = supp_D\left(X \cup Y \right) / supp_D\left( X \right)$. Support and confidence are two basic quality measures of association rules \cite{lenca2008selecting}. Most of association rules mining algorithms search for \textit{frequent} and \textit{confident} rules, that is for rules whose support and confidence are above or equal to a user-defined thresholds ($supp_D\left(X \rightarrow Y\right) \geq minSupp$ and $ conf_D\left(X\rightarrow Y\right) \geq minConf $). 

\subsection{Classification Rules and Contrast Patterns}

In many applications, the elements of the dataset are distributed among multiple classes. We restrict ourselves to the case of only two classes $Cl_1$ and $Cl_2$ in this work for the sake of simplicity. We also assume that the belonging of an element to a class is specified by the value of a particular attribute, which we refer to as \textit{class attribute} $Att_{Cl} = Cl_k$. If the consequent of an association rule is restricted to be composed of only the class attribute, then such rules are called \textit{classification rules} \cite{agrawal1994fast}. These rules can be used to address a very important task in class-labeled data: the task of identification of patterns that explain differences between classes of the dataset, or \textit{classification}.

 A group of patterns called \textit{contrast patterns} \cite{ramamohanarao2005efficient} was proposed as an alternative to classification rules. These patterns were also designed with the aim to understand differences between classes, however they are defined in a different way. An itemset is said to be a contrast pattern if it changes significantly its support from one class to another \cite{novak2009supervised}. This change can be measured in terms of, for example, \textit{growth rate}: $GrowthRate(X, Cl_1, Cl_2)=supp_{Cl_1}\left( X \right) / supp_{Cl_2}\left( X \right)$, where $supp_{Cl_k}\left( X \right)$ stands for support of itemset $X$ in class $Cl_k$. Given $\rho$ (with $\rho>1$)  a growth rate threshold, if $GrowthRate(X, Cl_1, Cl_2)\geq \rho$, then the itemset $X$ is said to be a $\rho$-emerging pattern from class $Cl_2$ to class $Cl_1$ \cite{dong1999efficient}.

Some works suggested that contrast patterns differ meaningfully from itemsets and thus association and classification rules \cite{ramamohanarao2005efficient}. However, it was shown in \cite{webb2003detecting} that a straightforward application of an existing association rules discovery algorithm can be successfully used to mine contrast patterns. It was also proven analytically in \cite{aleksandrova2017contrast} that classification rules and contrast patterns can be considered as equivalent. To be more precise, there exists a functional dependence between the value of the growth rate and the value of the confidence. For every given confidence threshold $\alpha$, it is possible to fix a growth rate threshold $\rho$ in such a way that all $\rho$-emerging patterns will also form confident classification rules with $minConf = \alpha$. 

\subsection{`Sets of Contrasting Rules'-Pattern}

Contrary to many state-of-the-art patterns, `Sets of Contrasting Rules'-pattern is composed of multiple rules and is based on notions of invariant and variable attributes \cite{aleksandrova2016setsecai,chertov2013fuzzy}.
An attribute is said to be \textit{varying} if its value can be changed externally to the system within the specified application task, and \textit{invariant} otherwise. For example, assume that we need to analyse medical data with an attribute $drug$ that specifies medicines taken by a patient, and another attribute $age$ corresponding to his age. If it is possible to change the medication then the attribute $drug$ is varying. At the same time, the value of the parameter $age$ cannot be changed. That is it belongs to the group of invariant attributes. The SCR-pattern is defined as follows.

\begin{definition}\label{def:SetOfContrastingRules}
For a parameter $\alpha$, a pair of classification rules $R1$ and $R2$ is called a pair of $\alpha$-contrasting rules if:
\end{definition}
\begin{enumerate}
	\item $conf(R1) \geq \alpha$ $\&$ $conf(R2) \geq \alpha$;
	\item $R1$ and $R2$ correspond to different classes;
	\item antecedents of the rules are made up of the same attributes; 
	\item the number of attributes is at least two and at least one of the attributes is varying;\label{defCond:cond4}
	\item values of all invariant attributes are the same for both rules;
	\item if there are no invariant attributes, then the values of at least one varying attribute is the same for both rules;\label{defCond:cond6}
	\item at least one varying attribute has different values in the pair.
\end{enumerate}

\begin{figure}
\begin{center}
\centerline{\includegraphics[scale=0.75]{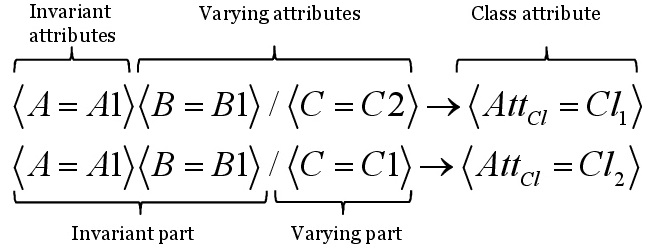} }
\caption{Example of an SCR-pattern\label{fig:SCRpattern}}
\end{center}
\end{figure}

We say that the rule $R2$ is a \textit{contrast pair} for the rule $R1$ and vice versa. An example of a pair of contrasting rules forming an SCR-pattern is given in Figure~\ref{fig:SCRpattern}. The antecedents of these rules contain one invariant attribute ($A$) and two varying attributes ($B$ and $C$). Sign `$/$' in the figure is used to visually separate invariant part of the pattern (attributes with same values) and its varying part (attributes with different values).

It was shown that, depending upon application, the varying part of the rule can be interpreted as \textit{local differences} between classes of the dataset \cite{aleksandrova2017contrast} or \textit{trigger factors} \cite{aleksandrova2016setsecai}. The term \textit{trigger factors} is used to reffer to the factors that can stimulate transfer of data elements from one class to another. It was also analytically proven that rules forming SCR-pattern are guaranteed to be of high-quality \cite{aleksandrova2017contrast}. This characteristic can be crucial for certain domains that require only highly confident knowledge (for example, in medicine).

For discovering SCR-patterns authors proposed to perform post-processing of association rules mined by classical algorithms \cite{aleksandrova2016setsecai}.  However, only a very small fraction of discovered rules have contrast pair and thus can be used. Thereby, despite its advantages `Sets of Contrasting Rules'-pattern has no efficient mining algorithm. In this paper we aim to solve this problem and to propose an efficient solution. 

\section{Mining Association Patterns}
Agrawal et al. in \cite{agrawal1993mining} introduced
the problem of association rules mining as \textit{the problem of mining all association rules that satisfy user-specified minimum support $minSupp$ and minimum confidence $minConf$ conditions}. This task is usually divided into two steps \cite{kotsiantis2006association}:
\begin{enumerate}
\item mining all frequent itemsets,  
\item forming confident rules of these itemsets.
\end{enumerate} 
Forming confident rules of frequent itemsets is straightforward and most of the algorithms concentrate on the first task. In contrast, the task of constructing all frequent itemsets is considered to be computationally expensive \cite{girotra2013comparative,hipp2000algorithms}. It is also known that its complexity grows exponentially with the number of attributes in the dataset. However, the search space of all possible itemsets can be effectively pruned basing on the downward-closure property: \textit{all subsets of a frequent itemset are themselves frequent itemsets} \cite{agrawal1993mining}. It means that every non-frequent itemset cannot be a part of any larger frequent itemset and it is possible to find a border that separates frequent and non-frequent itemsets. Thereby, the first task simply comes to searching for this border \cite{aggarwal2014frequent,hipp2000algorithms}.

 The search through the set of items can be done in depth-first or in width-first manners \cite{zaki1997new}. Both approaches result in the same set of frequent itemsets \cite{zaki2000scalable}. The width-first strategy forms the basis of multiple algorithms. Among them, there is Apriori algorithm proposed by Agrawal and Srikant \cite{agrawal1994fast}. The latter one is very popular and proved its efficiency in a great variety of applications.

\subsection{Apriori: Mining Frequent Itemsets}

\begin{figure}
\begin{center}
\centerline{\includegraphics[scale=0.60]{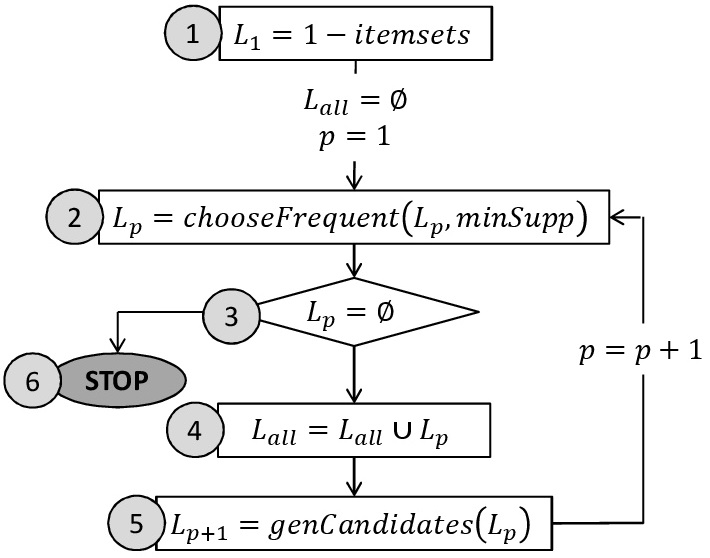} }
\caption{Apriori: mining frequent itemsets\label{fig:Apriori}}
\end{center}
\end{figure}

Let us  examine the procedure of constructing frequent itemsets used in Apriori algorithm (see Figure~\ref{fig:Apriori}). An itemset containing exactly $p$ attributes is referred to as \textit{$p$-itemset} and set $L_p$ stands for a set of all $p$-itemsets. The procedure starts with the initialisation of $L_1$ with the set of all possible 1-itemsets (all values of each attribute), see \textbf{step 1}.  After that, all non-frequent itemsets are excluded from $L_1$. This is done via comparing their support with a defined $minSupp$ value (corresponds to \textbf{step 2}). If $L_1$ contains frequent itemsets (\textbf{step 3}), then these itemsets are added to the list of all frequent itemsets $L_{all}$ (\textbf{step 4}). Finally, the set of all 2-itemsets $L_2$ is constructed on the \textbf{step 5}. From the downward-closure property, it follows that a 2-itemset is not frequent if at least one its subsets does not belong to the set of frequent 1-itemsets ($L_1$). Thereby, the set $L_2$ can be constructed from the set $L_1$ through the operation of self-joining (see \cite{agrawal1994fast} for more details). This allows reducing the number of 2-itemset to be considered by eliminating all those, that are known to be non-frequent. Afterwards, the procedure restarts from the step 2 processing the set $L_2$ in a similar way. The procedure finishes when a newly generated on step~5 and filtered on step~2 set $L_p$  is empty. The resulting set $L_{all}$ consists of all frequent itemsets of different sizes.

Finally, all possible confident association rules are constructed by exploiting the set of generated frequent itemsets. Suppose we have a frequent itemset $c$ and its non-empty subset $\beta$. Then an association rule is constructed as follows: $\left(c-\beta\right) \rightarrow \beta $. If the resulting rule meets the $minConf$ condition it is added to the list of confident rules.

\subsection{CAR-Apriori: Mining Frequent Ruleitems}

Apriori algorithm can be used for the identification of classification rules through post-processing of the obtained association rules: we choose only those rules, whose consequents are composed of only  class attribute. However,  computational costs can be reduced if the search of classification rules is incorporated into  Apriori algorithm. Such an algorithm, CAR-Apriori, was proposed in \cite{ma1998integrating}. This algorithm is based on a new concept: \textit{ruleitem}. A ruleitem is a construction of the form $condset<Att_{Cl}=Cl_k>$. A $condset$ is essentially an itemset, and expression $<Att_{Cl}=Cl_k>$ specifies the value of the class attribute. The difference between a condset and an itemset is that the latter one can contain class attributes $Att_{Cl}$ while the former one no.

The support of a ruleitem is defined as a ratio of the number of elements containing the specified condset and belonging to the class $Cl_k$ to the number of all elements in the dataset. Each ruleitem essentially represents a rule of the form $condset \rightarrow Att_{Cl}=Cl_k $ with the support equal to the support of the ruleitem and the confidence calculated as a fraction of the support of the ruleitem to the support of the corresponding condset.

The algorithm of mining classification rules is the same as Apriori with the difference that instead of frequent itemsets frequent ruleitems are mined. Also, classification rules are formed directly from ruleitems. Both Apriori and CAR-Apriori result in the same set of classification rules. However, the incorporation of class-specific information that is done in CAR-Apriori allows to speed-up the process.

\section{SCR-Apriori: Mining Frequent and Contrast SCR-Ruleitems}

\subsection{Intuition}

Similar to using Apriori for mining classification rules, it is possible to use CAR-Apriori for mining `Sets of Contrasting Rules'-pattern via filtering out all classification rules that have no contrast pair. However, this solution is not efficient. 

We propose to further extend the idea of CAR-Apriori algorithm for direct mining of ruleitems that can potentially form SCR-pattern. Apart of excluding all non-frequent itemsets (as it is done in Apriori) and all itemsets that cannot form classification rules (as it is done in CAR-Apriori), we propose to further exclude those itemsets that cannot be used for constructing SCR-patterns, even if they can form frequent and confident classification rules on their own. 

 Recall that an SCR-pattern consists of two contrasting rules. Thereby, if for a certain frequent ruleitem there is no other frequent ruleitem that can potentially form a pair of contrast classification rules with the first ruleitem, then it can be excluded from the search space regardless of its frequency. In order to fulfil this task, we propose a new algorithm SCR-Apriori. This algorithm aims at mining all frequent and contrast \textit{SCR-ruleitems} and forms SCR-patterns of them.

\subsection{SCR-Apriori Algorithm}

We define a \textit{SCR-ruleitem} as follows: $condset{<\!\!supp_{Cl_1},supp_{Cl_2}\!\!>}$, where $supp_{Cl_k}$ stands for support of the condset on the class $Cl_k$. Essentially an SCR-ruleitem is a condset accompanied with its support values in all classes defined on the dataset $D$. We also define a \textit{contrast pair} for an SCR-ruleitem in a similar way as we defined a contrast pair for a rule (see Definition \ref{def:SetOfContrastingRules}) with the following differences:
\begin{itemize}
\item SCR-ruleitems in the pair can be composed of only one attribute that can be either invariant or varying (compare with condition \ref{defCond:cond4} in Definition \ref{def:SetOfContrastingRules});
\item all varying attributes can have different values even if there are no invariant attributes in the condsets of the considered SCR-ruleitems (compare with condition \ref{defCond:cond6}).
\end{itemize}

\begin{figure}
\begin{center}
\centerline{\includegraphics[scale=0.6]{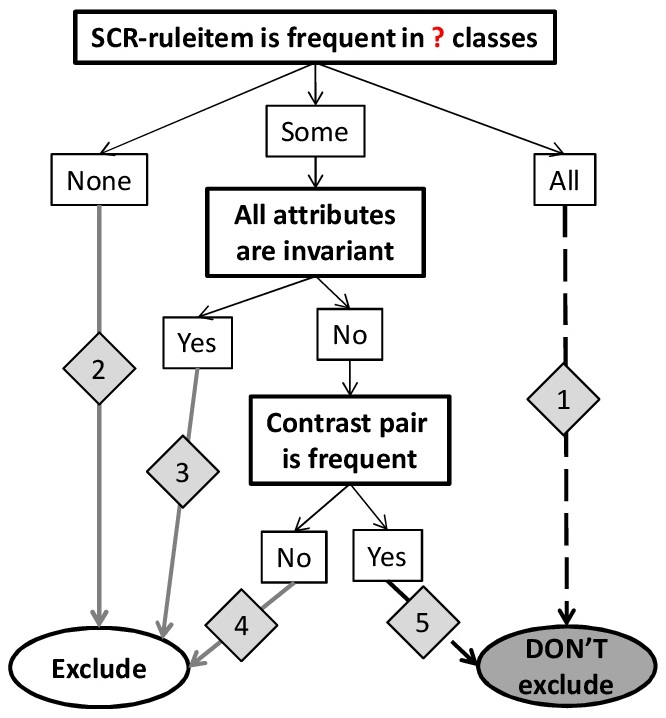} }
\caption{Function for choosing frequent and contrast SCR-ruleitems\label{fig:MyAlgo}}
\end{center}
\end{figure}

The procedure of mining frequent and contrast SCR-ruleitems is very similar to the procedure of mining frequent itemsets and frequent ruleitems presented in Figure~\ref{fig:Apriori}. The only difference is in \textbf{step 2}. On this step SCR-ruleitems are filtered out from the set $L_p$ not only depending on the value of their support on different classes but also on the fact if a given SCR-ruleitem has a \textit{contrast pair} frequent on the opposite class. Considering this, the function $chooseFrequent$ in the algorithm in Figure~\ref{fig:Apriori} should be replaced with the function $chooseFrequentAndContrast$. This function decides whether to exclude or not an SCR-ruleitem from $L_p$ according to the algorithm given in Figure~\ref{fig:MyAlgo}.
 
\begin{figure}
\begin{center}
\centerline{\includegraphics[scale=0.5]{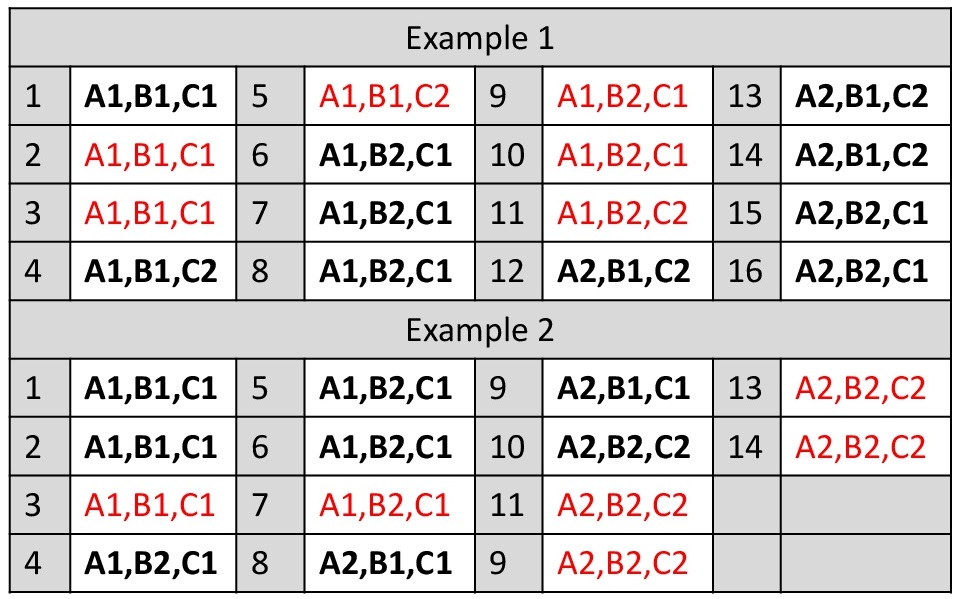} }
\caption{List of records for Example 1 and Example 2\label{fig:Transactions}}
\end{center}
\end{figure}

We analyse the structure and visualise different branches of the proposed algorithm using two examples. We utilize 2 sets of records consisting of 16 and 14 elements respectively (see Figure~\ref{fig:Transactions}). All records are defined on 3 attributes $A$, $B$, and $C$. Each of these attributes has 2 possible values which we distinguish by numbers, e.g. the possible values of the first attribute are $A1$ and $A2$. Among these 3 attributes the first one ($A$) is invariant and the rest of the attributes ($B$ and $C$) are varying. There are 2 classes $Cl_1$ and $Cl_2$ defined on each set of records. Records belonging to the first class are in \textbf{bold} while records belonging to the second class are {\color{BrickRed}wine-red}.

\begin{figure*}
\begin{center}
\centerline{\includegraphics[scale=0.6]{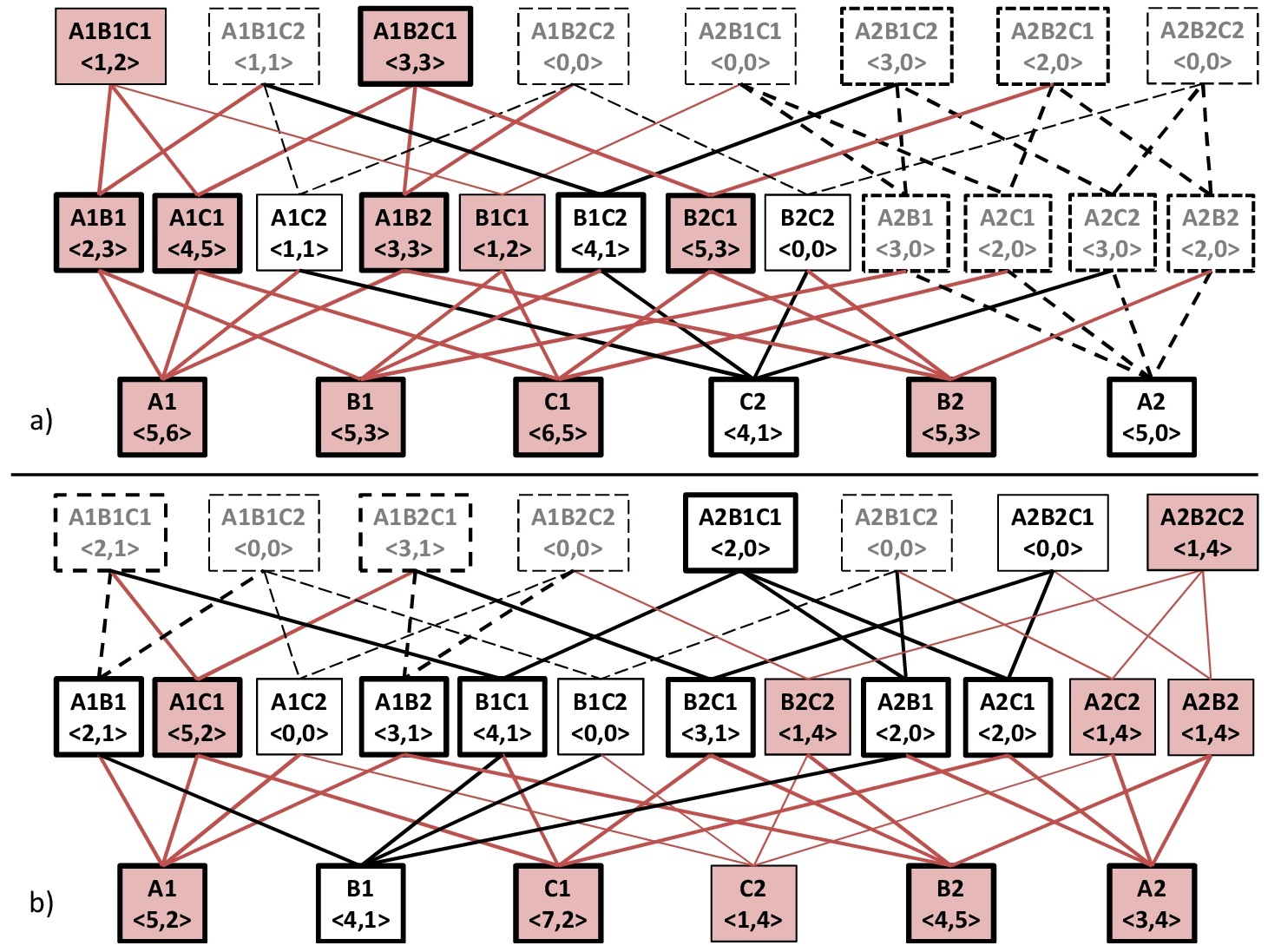} }
\caption{Contrast ruleitems for Example 1 (a) and Example 2 (b) \label{fig:ContrastRuleItems}}
\end{center}
\end{figure*}

Figures \ref{fig:ContrastRuleItems}a and \ref{fig:ContrastRuleItems}b show all SCR-ruleitems for the first (a) and the second (b) examples respectively. Each square represents a particular SCR-ruleitem. For the sake of simplicity we present not the support of a condset on each class, but its support number. For both examples, we fix the minimum support number threshold equal to 2.

The fact that a particular condset is frequent in $Cl_1$ or $Cl_2$ is visualised with a think border or a wine-red background color respectively. A particular SCR-ruleitem having both a thick border and a wine-red background is frequent in both classes. A dashed border means that a particular SCR-ruleitem was not constructed by our algorithm due to exclusion of one or more of its subsets by by our algorithm.

The shape of links between SCR-ruleitems also reflects their frequency. In particular, a link up-coming from an SCR-ruleitem is thick or/and of wine-red color, if this SCR-ruleitem is frequent on the first or/and on the second class.  Additionally, an up-coming link is dashed if the SCR-ruleitem was marked by our algorithm as the one that cannot form an SCR-pattern.

Let us now analyse the algorithm given in Figure~\ref{fig:MyAlgo}. First, this algorithm examines the values of support of an SCR-ruleitem in all classes defined on the dataset. 

\textbf{Branch 1}. If it is frequent on all classes then it can form an SCR-pattern even if it has no frequent contrast pair. Indeed, let us have a look on SCR-ruleitem $A1C1{<\!\!4,5\!\!>}$ in Figure~\ref{fig:ContrastRuleItems}-a. There is only one SCR-ruleitem that can form a contrast pair for it: $A1C2{<\!\!1,1\!\!>}$ and it is not frequent on any of 2 classes. However, we can form SCR-ruleitems $A1B1C1{<\!\!1,2\!\!>}$ and $A1B2C1{<\!\!3,3\!\!>}$ which will result in SCR-pattern $\{A1C1/B1 \rightarrow Att_{Cl}=Cl_2, conf=0.67 \; : \; A1C1/B2 \rightarrow Att_{Cl}=Cl_1, conf=0.5\}$. Thereby, an SCR-ruleitem that is frequent on all classes should not be excluded.

\textbf{Branch 2} of the algorithm corresponds to the case when the SCR-ruleitem is not frequent on any of the classes and suggests to exclude it. Indeed, if the condset of the SCR-ruleitem is not frequent, then no frequent rules can be formed neither from this condset itself, nor from its super-sets. And, thereby, no SCR-patterns can be produced (see for example SCR-ruleitems $A1C2{<\!\!1,1\!\!>}$ and $B2C2{<\!\!0,0\!\!>}$ in Figure~\ref{fig:ContrastRuleItems}a or $A1C2{<\!\!0,0\!\!>}$ and $B1C2{<\!\!0,0\!\!>}$ in Figure~\ref{fig:ContrastRuleItems}b).

If the SCR-ruleitem is frequent on only a fraction of classes, then the nature of its attributes should be considered. 

\textbf{Branch 3.} Let us consider the case when all attributes forming SCR-ruleitem are invariant, for example SCR-ruleitem $A2{<\!\!5,0\!\!>}$ in Figure~\ref{fig:ContrastRuleItems}a. Both rules from an SCR-pattern should have the same values of all invariant attributes. It means that it is impossible to construct an SCR-pattern in this case as some components of the rules forming this SCR-pattern will be frequent on only one class. In our example $A2$ is frequent only in the first class.

Finally, if the condset of the SCR-ruleitem consists of either varying or both varying and invariant attributes then the frequency of its contrast pairs on the opposite class is checked.

\textbf{Branch 4} considers the case when SCR-ruleitem is frequent in one class but has no contrast pair frequent on another class. Let us consider as an example SCR-ruleitem $A1B1{<\!\!2,1\!\!>}$ in Figure~\ref{fig:ContrastRuleItems}b. The only contrast pair for this SCR-ruleitem is $A1B2{<\!\!3,1\!\!>}$. No SCR-patterns can be produced of this pair as no frequent classification rule can be formed for the second class. From the downward-closure property it also follows that all supersets of $A1B1{<\!\!2,1\!\!>}$ will not be frequent on the second class as well, for example, $A1B1C1{<\!\!2,1\!\!>}$. Thereby, supersets of the considered SCR-ruleitem can potentially form an SCR-pattern only with  those SCR-ruleitems that have different values of at least one invariant attribute from the considered SCR-ruleitem. They have to be also frequent on the other class. However, each of such SCR-ruleitems has a subset that forms a contrast pair with the considered SCR-ruleitem. But we know that all of them are non-frequent on the other class. This means that no SCR-patterns can be formed from the supersets as well. In the considered example, SCR-ruleitem $A1B1{<\!\!2,1\!\!>}$ has only one invariant attribute $B$ with the value $B1$. Thereby, $A1B1C1{<\!\!2,1\!\!>}$ as a superset of $A1B1{<\!\!2,1\!\!>}$ can potentially form SCR-pattern with either $A1B2C1{<\!\!3,1\!\!>}$ or $A1B2C2{<\!\!0,0\!\!>}$. However, as we know that $A1B2{<\!\!3,1\!\!>}$ is not frequent on the second class, we can conclude the same about its supersets without calculating their support explicitly. The same reasoning is valid for another superset of $A1B1{<\!\!2,1\!\!>}$, namely for $A1B1C2{<\!\!0,0\!\!>}$.

Finally, \textbf{branch 5} of our algorithm considers the case of partially frequent SCR-ruleitems that have contrast pairs. We can show on an example that in this case it is possible to construct SCR-patterns and thus such SCR-ruleitems should not be excluded. Let us consider $B1C1{<\!\!4,1\!\!>}$ in Figure~\ref{fig:ContrastRuleItems}b which is frequent on the first class. Its contrast pair $B2C2{<\!\!1,4\!\!>}$ is frequent on the second class. Using the supersets of these SCR-ruleitems we can construct SCR-pattern $\{A2/B1C1 \rightarrow Att_{Cl}=Cl_1, conf=1 : A2/B2C2 \rightarrow Att_{Cl}=Cl_2, conf=0.8\}$. Note, that SCR-ruleitem $C2{<\!\!4,1\!\!>}$ in Figure~\ref{fig:ContrastRuleItems}a is considered in a similar way. It has a contrast pair $C1{<\!\!6,5\!\!>}$ that is frequent on the second class. And it is also possible to construct SCR-pattern of their supersets $B1C1{<\!\!1,2\!\!>}$ and $B1C2{<\!\!4,1\!\!>}$.

Using the algorithm given in Figure~\ref{fig:MyAlgo} it is possible to perform similar analysis for all SCR-ruleitems in Figure~\ref{fig:ContrastRuleItems} and choose those that satisfy conditions of frequency and contrast. Note that, for example, SCR-ruleitem $A2B1C2{<\!\!3,0\!\!>}$ in Figure~\ref{fig:ContrastRuleItems}a is not considered by our algorithm even though it can form a frequent and confident association rule. No SCR-patterns can be formed of this SCR-ruleitem and it is excluded from the consideration because its subset $A2$ does not satisfy the conditions of our algorithm (branch 3).

When all frequent and contrast SCR-ruleitems are discovered, we can construct SCR-patterns directly using SCR-ruleitems and their contrast pairs. Note, however, that not all constructed SCR-ruleitems will form SCR-patterns. For example, if the parameter $\alpha$ is set to $\alpha = 0.6$, no SCR-patterns can be produced from a pair of SCR-ruleitems $A1B1C1{<\!\!1,2\!\!>}$ and $A1B2C1{<\!\!3,3\!\!>}$ in Figure~\ref{fig:ContrastRuleItems}a.

\subsection{Performance evaluation}

In order to quantify the gain in performance of SCR-Apriori as compared to CAR-Apriori we used the same dataset as in state-of-the-art works \cite{aleksandrova2016setsecai,aleksandrova2016sets,chertov2013fuzzy}: 5-percent sample of the California census dataset for the year 2000\footnote{https://www.census.gov/prod/cen2000/doc/pums.pdf}.
We also performed the same preprocessing steps as in the cited papers. After that, we generated SCR-patterns with SCR-Apriori and via post-filtering of classification rules obtained with CAR-Apriori, as suggested in the state-of-the-art.

 We mined SCR-patterns with $minConf = 0.5$ and $minSupp = 0.07$. The value of support is chosen to be so small in accordance with previous works \cite{aleksandrova2016sets,chertov2013fuzzy} and with the statement that c\textit{ontrast patterns with small support can be of particular interest} \cite{chertov2013fuzzy}.
 
  We obtained the same set of SCR-patterns with both approaches. However, the number of mined SCR-ruleitems corresponds to 58\% of the number of frequent ruleitems identified with CAR-Apriori. Also, the number of rules generated by SCR-Apriori corresponds to only 6.4\% of the number of classification rules obtained with the state-of-the-art procedure. Thereby, the proposed algorithm allows obtaining the same results but with the substantial decrease of computation costs. 

\section{Conclusions}
In this paper, we proposed a novel SCR-Appriori algorithm for direct mining of `Sets of Contrasting Rules'-pattern. Through experimental results on real dataset, we showed that SCR-Apriori allows decreasing significantly computational costs, as compared to the state-of-the-art approach.

The proposed algorithm incorporates the knowledge about the structure of SCR-pattern into the mining process and is essentially a modified version of known Apriori and CAR-Apriori algorithms.

\bibliographystyle{IEEEtran}
\bibliography{AleksandrovaChertov}

\begin{thebibliography}{10}
\providecommand{\url}[1]{#1}
\csname url@samestyle\endcsname
\providecommand{\newblock}{\relax}
\providecommand{\bibinfo}[2]{#2}
\providecommand{\BIBentrySTDinterwordspacing}{\spaceskip=0pt\relax}
\providecommand{\BIBentryALTinterwordstretchfactor}{4}
\providecommand{\BIBentryALTinterwordspacing}{\spaceskip=\fontdimen2\font plus
\BIBentryALTinterwordstretchfactor\fontdimen3\font minus
  \fontdimen4\font\relax}
\providecommand{\BIBforeignlanguage}[2]{{%
\expandafter\ifx\csname l@#1\endcsname\relax
\typeout{** WARNING: IEEEtran.bst: No hyphenation pattern has been}%
\typeout{** loaded for the language `#1'. Using the pattern for}%
\typeout{** the default language instead.}%
\else
\language=\csname l@#1\endcsname
\fi
#2}}
\providecommand{\BIBdecl}{\relax}
\BIBdecl

\bibitem{kotsiantis2006association}
S.~Kotsiantis and D.~Kanellopoulos, ``Association rules mining: A recent
  overview,'' \emph{GESTS International Transactions on Computer Science and
  Engineering}, vol.~32, no.~1, pp. 71--82, 2006.

\bibitem{lenca2008selecting}
P.~Lenca, P.~Meyer, B.~Vaillant, and S.~Lallich, ``On selecting interestingness
  measures for association rules: User oriented description and multiple
  criteria decision aid,'' \emph{European journal of operational research},
  vol. 184, no.~2, pp. 610--626, 2008.

\bibitem{techapichetvanich2004visual}
K.~Techapichetvanich and A.~Datta, ``Visual mining of market basket association
  rules,'' in \emph{Computational Science and Its Applications--ICCSA
  2004}.\hskip 1em plus 0.5em minus 0.4em\relax Springer, 2004, pp. 479--488.

\bibitem{aleksandrova2016setsecai}
M.~Aleksandrova, A.~Brun, O.~Chertov, and A.~Boyer, ``Sets of contrasting rules
  to identify trigger factors,'' in \emph{ECAI 2016: 22nd European Conference
  on Artificial Intelligence}.\hskip 1em plus 0.5em minus 0.4em\relax IOS
  Press, 2016.

\bibitem{aleksandrova2017contrast}
M.~Aleksandrova, O.~Chertov, A.~Brun, and A.~Boyer, ``Contrast classification
  rules for mining local differences in medical data,'' in \emph{Intelligent
  Data Acquisition and Advanced Computing Systems: Technology and Applications
  (IDAACS), 2017 9th IEEE International Conference on}, vol.~2.\hskip 1em plus
  0.5em minus 0.4em\relax IEEE, 2017, pp. 880--883.

\bibitem{aleksandrova2016sets}
M.~Aleksandrova, A.~Brun, O.~Chertov, and A.~Boyer, ``Sets of contrasting
  rules: A supervised descriptive rule induction pattern for identification of
  trigger factors,'' in \emph{Tools with Artificial Intelligence (ICTAI), 2016
  IEEE 28th International Conference on}.\hskip 1em plus 0.5em minus
  0.4em\relax IEEE, 2016, pp. 431--435.

\bibitem{agrawal1993mining}
R.~Agrawal, T.~Imieli{\'n}ski, and A.~Swami, ``Mining association rules between
  sets of items in large databases,'' \emph{ACM SIGMOD Record}, vol.~22, no.~2,
  pp. 207--216, 1993.

\bibitem{ma1998integrating}
B.~Liu, W.~Hsu, and M.~Yiming, ``Integrating classification and association
  rule mining,'' in \emph{Proceedings of the fourth international conference on
  knowledge discovery and data mining}, 1998, pp. 80--86.

\bibitem{agrawal1994fast}
R.~Agrawal, R.~Srikant \emph{et~al.}, ``Fast algorithms for mining association
  rules,'' in \emph{Proc. 20th int. conf. very large data bases, VLDB}, vol.
  1215, 1994, pp. 487--499.

\bibitem{ramamohanarao2005efficient}
K.~Ramamohanarao, J.~Bailey, and H.~Fan, ``Efficient mining of contrast
  patterns and their applications to classification,'' in \emph{2005 3rd
  International Conference on Intelligent Sensing and Information
  Processing}.\hskip 1em plus 0.5em minus 0.4em\relax IEEE, 2005, pp. 39--47.

\bibitem{novak2009supervised}
P.~K. Novak, N.~Lavra{\v{c}}, and G.~I. Webb, ``Supervised descriptive rule
  discovery: A unifying survey of contrast set, emerging pattern and subgroup
  mining,'' \emph{The Journal of Machine Learning Research}, vol.~10, pp.
  377--403, 2009.

\bibitem{dong1999efficient}
G.~Dong and J.~Li, ``Efficient mining of emerging patterns: Discovering trends
  and differences,'' in \emph{Proceedings of the fifth ACM SIGKDD international
  conference on Knowledge discovery and data mining}.\hskip 1em plus 0.5em
  minus 0.4em\relax ACM, 1999, pp. 43--52.

\bibitem{webb2003detecting}
G.~I. Webb, S.~Butler, and D.~Newlands, ``On detecting differences between
  groups,'' in \emph{Proceedings of the ninth ACM SIGKDD international
  conference on Knowledge discovery and data mining}.\hskip 1em plus 0.5em
  minus 0.4em\relax ACM, 2003, pp. 256--265.

\bibitem{chertov2013fuzzy}
O.~Chertov and M.~Aleksandrova, ``Fuzzy clustering with prototype extraction
  for census data analysis,'' in \emph{Soft Computing: State of the Art Theory
  and Novel Applications}.\hskip 1em plus 0.5em minus 0.4em\relax Springer,
  2013, pp. 289--313.

\bibitem{girotra2013comparative}
M.~Girotra, K.~Nagpal, S.~Minocha, and N.~Sharma, ``Comparative survey on
  association rule mining algorithms,'' \emph{International Journal of Computer
  Applications}, vol.~84, no.~10, 2013.

\bibitem{hipp2000algorithms}
J.~Hipp, U.~G{\"u}ntzer, and G.~Nakhaeizadeh, ``Algorithms for association rule
  mining — a general survey and comparison,'' \emph{ACM sigkdd explorations
  newsletter}, vol.~2, no.~1, pp. 58--64, 2000.

\bibitem{aggarwal2014frequent}
C.~C. Aggarwal, M.~A. Bhuiyan, and M.~Al~Hasan, ``Frequent pattern mining
  algorithms: A survey,'' in \emph{Frequent pattern mining}.\hskip 1em plus
  0.5em minus 0.4em\relax Springer, 2014, pp. 19--64.

\bibitem{zaki1997new}
M.~J. Zaki, S.~Parthasarathy, M.~Ogihara, W.~Li \emph{et~al.}, ``New algorithms
  for fast discovery of association rules,'' in \emph{KDD}, vol.~97, 1997, pp.
  283--286.

\bibitem{zaki2000scalable}
M.~J. Zaki, ``Scalable algorithms for association mining,'' \emph{IEEE
  Transactions on Knowledge and Data Engineering}, vol.~12, no.~3, pp.
  372--390, 2000.

\end{thebibliography}
\end{document}